\documentclass[10pt,twocolumn,letterpaper]{article}

\usepackage{cvpr}              

\usepackage{multirow}
\usepackage[accsupp]{axessibility} 

\definecolor{cvprblue}{rgb}{0.21,0.49,0.74}
\usepackage[pagebackref,breaklinks,colorlinks,allcolors=cvprblue]{hyperref}

\def\paperID{4114} 
\def\confName{CVPR}
\def\confYear{2025}

\title{MUST: The First Dataset and Unified Framework for \\ Multispectral UAV Single Object Tracking}

\author{Haolin Qin, Tingfa Xu$^{\dagger}$, Tianhao Li, Zhenxiang Chen, Tao Feng, Jianan Li$^{\dagger}$\\
Beijing Institute of Technology\\
{\tt\small \{3120225333,ciom\underline{ }xtf1,lijianan\}@bit.edu.cn}
\thanks{$^{\dagger}$ Correspondence to: Tingfa Xu and Jianan Li.}
}

\begin{document}
\maketitle
\begin{abstract}
UAV tracking faces significant challenges in real-world scenarios, such as small-size targets and occlusions, which limit the performance of RGB-based trackers. Multispectral images (MSI), which capture additional spectral information, offer a promising solution to these challenges. However, progress in this field has been hindered by the lack of relevant datasets. To address this gap, we introduce the first large-scale Multispectral UAV Single Object Tracking dataset (MUST), which includes 250 video sequences spanning diverse environments and challenges, providing a comprehensive data foundation for multispectral UAV tracking. We also propose a novel tracking framework, UNTrack, which encodes unified spectral, spatial, and temporal features from spectrum prompts, initial templates, and sequential searches. UNTrack employs an asymmetric transformer with a spectral background eliminate mechanism for optimal relationship modeling and an encoder that continuously updates the spectrum prompt to refine tracking, improving both accuracy and efficiency. Extensive experiments show that our proposed UNTrack outperforms state-of-the-art UAV trackers. We believe our dataset and framework will drive future research in this area. The dataset is available on https://github.com/q2479036243/MUST-Multispectral-UAV-Single-Object-Tracking.
\end{abstract}    
\section{Introduction}
\label{sec:intro}

\begin{figure}[t]
  \centering
  \includegraphics[width=1.0\linewidth]{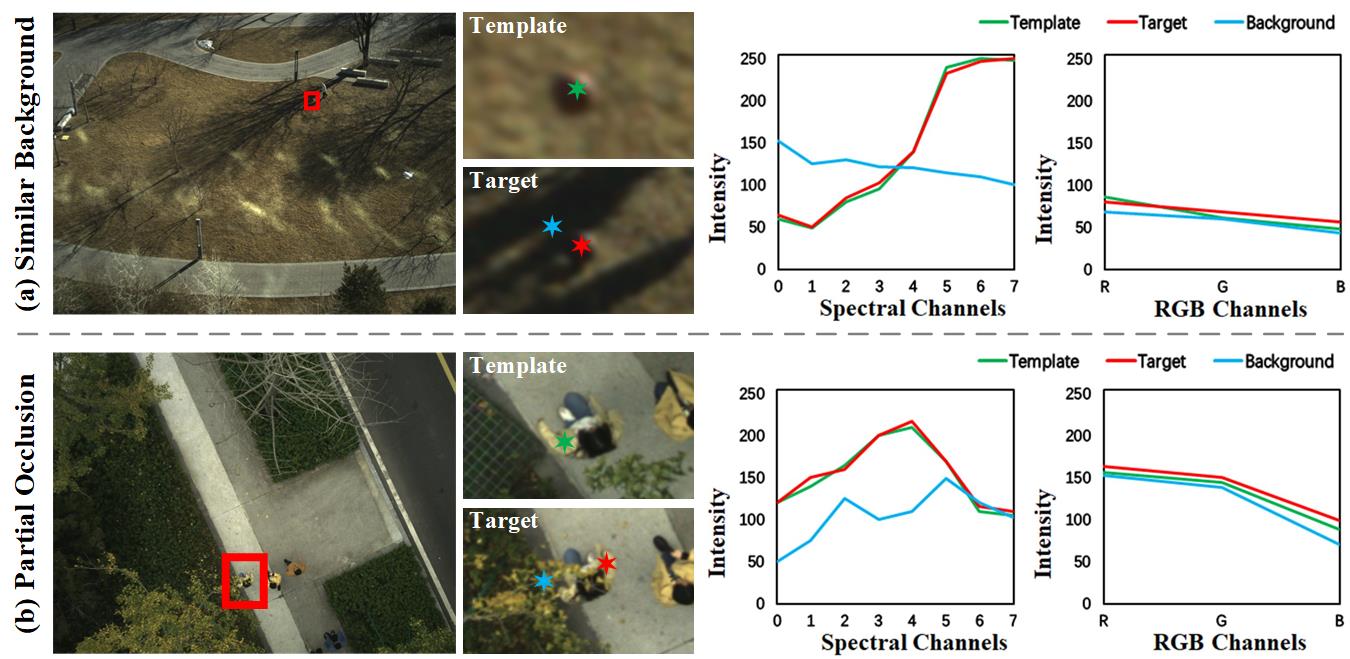}
  \caption{In challenging scenarios, the spatial features (\eg, color and texture) of the tracked target closely resemble those of the background, making differentiation and localization difficult. However, the target’s spectral information differs significantly from the background and aligns with the template’s spectral data, providing robust features for reliable tracking.}
  \label{fig:motivation}
\end{figure}

UAV tracking is critical but often encounters complex challenges, which hinder the extraction of accurate spatial features, as shown in \cref{fig:motivation}. These factors substantially degrade the performance of RGB-based trackers that rely primarily on visual cues~\cite{8060595, Li_2017_CVPR}. Furthermore, the target's appearance may change significantly during tracking, often deviating considerably from the initial template, thereby making the tracker prone to drift.

Multispectral images provide additional spectral information that characterizes the target's intrinsic reflectance properties~\cite{qin2024dmssn}, offering valuable insights in situations where spatial features are limited by the aforementioned challenges. For example, even with severe background interference and color similarities, distinct spectral curves allow for effective target differentiation and localization (\cref{fig:motivation}). Moreover, key spectral features tend to remain stable during tracking, improving robustness. Thus, multispectral imaging is a promising solution for UAV tracking in challenging scenarios. However, the lack of large-scale, challenging datasets hampers progress in this area.

To address this gap, we present the first large-scale Multispectral UAV Single Object Tracking dataset (MUST). The dataset comprises 250 video sequences, totaling 43,000 frames, each with 1200$\times$900 pixels and 8 spectral bands that encompass both visible-light and near-infrared bands. Furthermore, the dataset incorporates 12 key challenge attributes, including small-size targets, background clutter, and similar colors, making it highly representative of real-world challenges. By leveraging data diversity and including difficult scenarios, the MUST dataset provides crucial support for the development and thorough evaluation of multispectral UAV tracking methods.

In addition, we introduce a novel tracking framework, termed the Unified Spectral-Spatial-Temporal Tracker (UNTrack), designed specifically for multispectral UAV tracking. UNTrack simultaneously integrates spectral, spatial, and temporal information in a unified manner by incorporating the historical spectrum prompt, the initial template, and sequential searches from consecutive frames. To better extract collaborative features from the spectral, spatial, and temporal domains, UNTrack employs an asymmetric transformer that models the core relationships among various feature domains while pruning attention maps irrelevant to tracking. This innovative asymmetric structure streamlines and compacts joint feature extraction, striking an optimal balance between accuracy and efficiency.

To further reduce noisy interference and computational costs, UNTrack employs a background suppression mechanism that efficiently eliminates background regions from the search image based on spectral characteristics. Additionally, UNTrack incorporates an encoder that generates a spectral prompt that captures the unique spectral features of the target, enabling effective discrimination between the target and similar background. The encoded spectral prompt is then used to update the input for subsequent tracking, providing stable spectral information that is free from external interference, thereby enhancing tracking robustness.

We conducted a thorough evaluation of the proposed UNTrack and existing trackers using our newly introduced MUST dataset. UNTrack demonstrates state-of-the-art performance, and extended experiments on other multispectral or general object tracking datasets highlight its superiority and versatility. In summary, the main contributions of this work are as follows:

\begin{itemize}
\item We present MUST, the first large-scale and challenging multispectral UAV tracking dataset, providing foundational data support to advance research in this field.

\item We propose a novel framework, UNTrack, that integrates spectral, spatial, and temporal information in a unified manner for robust tracking in challenging scenarios.

\item We introduce an asymmetric transformer that models the core relationships among various
feature domains, pruning attention maps and improving tracking efficiency.
\end{itemize}
\section{Related Work}
\label{sec:related}

\noindent\textbf{UAV Tracking Datasets.}
With the growing interest in UAV tracking, numerous related datasets have emerged. Examples include Stanford Drone~\cite{robicquet2016learning}, UAV123~\cite{mueller2016benchmark}, and others~\cite{li2017visual, zhu2021detection, 10325629}, which provide diverse data to the field. However, these RGB-based datasets struggle with the challenges of obvious appearance variations in applications, limiting their effectiveness in evolving tracking needs. To address this limitation, Xiong \etal~\cite{xiong2020material} introduced the first MSI-based tracking dataset, but it focuses on general objects thereby unsuitable for UAV tracking. To our knowledge, there are currently no multispectral datasets specifically designed for UAV tracking. In response, we propose a large-scale dataset with a wide range of challenging attributes, addressing the issue of data scarcity.

\noindent\textbf{Multispectral UAV Trackers.}
In general visual tracking, one-stream trackers~\cite{ye2022joint} integrate feature extraction and interaction within a single framework, offering superior performance compared to Siamese trackers~\cite{li2018high} and becoming the mainstream approach. However, existing UAV trackers are still primarily based on the Siamese architecture~\cite{li2024learning}, while only focusing on RGB data and ignoring more potential multispectral images. Furthermore, previous studies~\cite{li2023learning, li2024material} mostly adapted RGB-based trackers to handle multispectral images without fully exploiting spectral features. To address these issues, we propose the first multispectral UAV tracker, equipped with a novel tracking framework, to tackle challenges based on unified features.

\noindent\textbf{Prompt Learning.}
Recent studies have introduced prompts to improve tracking performance~\cite{zheng2024odtrack, bai2024artrackv2}, but these approaches focus primarily on appearance, overlooking the more fundamental spectral characteristics. Moreover, previous trackers~\cite{song2023compact, cui2022mixformer} have not explored the optimal attention structure for integrating prompts. The unified extraction of spectral, spatial, and temporal features remains a critical challenge. To address this, we propose an asymmetric transformer and a spectrum prompt encoder that strike a balance between tracking accuracy and inference efficiency.

\section{MUST Dataset}
\label{sec:dataset}

\subsection{Overview}

\begin{table*}
  \caption{Comparison with other UAV perspective datasets and MSI-based datasets. $\textendash$ means no relevant information was provided.}
  \label{tab:comparison}
  \small
  \centering
  \setlength{\tabcolsep}{6pt}
  \begin{tabular}{lccccccccc}
    \toprule
    Dataset & Modality & Task & Sequences & Frames & Resolution & Channels & Durations & Altitude & Year\\ 
    \midrule  
    UAV123~\cite{mueller2016benchmark} & RGB & UAV SOT & 123 & 113 k & 1280$\times$720 & 3 & 
    3800 s & 5$\sim$25 m & 2016 \\

    DTB70~\cite{li2017visual} & RGB & UAV SOT & 70 & 16 k  & 1280$\times$720 & 3 & 640 s & 0$\sim$120 m & 2017 \\

    VisDrone~\cite{zhu2021detection} & RGB & UAV SOT & 167 & 139 k& Not Uniform & 3 & 4600 s  & \textendash & 2021 \\

    VTUAV~\cite{zhang2022visible} & RGB-T & UAV SOT &500 & 1700 k & 1920$\times$1080 & 4 & 55000 s & 5$\sim$20 m & 2022 \\
    HOT~\cite{xiong2020material} & MSI & General SOT & 50 & 21 k & 512$\times$256 & 16 & 840 s & \textendash & 2020 \\
    
    \midrule 
    MUST (Ours) & MSI & UAV SOT & 250 & 43 k & 1200$\times$900 & 8 & 8600 s & 20$\sim$250 m & 2024 \\
    \bottomrule
  \end{tabular}
\end{table*}

\noindent \textbf{Data Collection.} The proposed Multispectral UAV Single Object Tracking (MUST) dataset is captured using a professional-grade UAV equipped with a multispectral camera. To enhance data diversity, we collected data across various scenarios and time periods, covering 250 video sequences, of which 160 are used for training and 90 are used for testing. Each multispectral image frame has a spatial resolution of $1200\times900$ and 8 spectral bands. A detailed description of the dataset parameters and processing strategies is provided in the supplementary materials.

\noindent \textbf{Dataset Comparison.} \Cref{tab:comparison} shows the comparison of our proposed MUST dataset against other UAV or multispectral tracking datasets. Although RGB-based datasets~\cite{mueller2016benchmark, li2017visual} are well-established, their single data modality limits the use of spectral information. Furthermore, the existing MSI-based tracking dataset~\cite{xiong2020material} is limited by its small-scale and artificial ground scenarios, restricting its applicability to general tracking. In contrast, our proposed MUST dataset offers a larger number of video sequences with extended durations, ensuring both data diversity and broad challenges. In addition, we collected UAV data across a broader range of flight altitudes with various flight postures, making it more reflective of real-world scenarios.

\begin{figure}[t]
  \centering
  \includegraphics[width=1\linewidth]{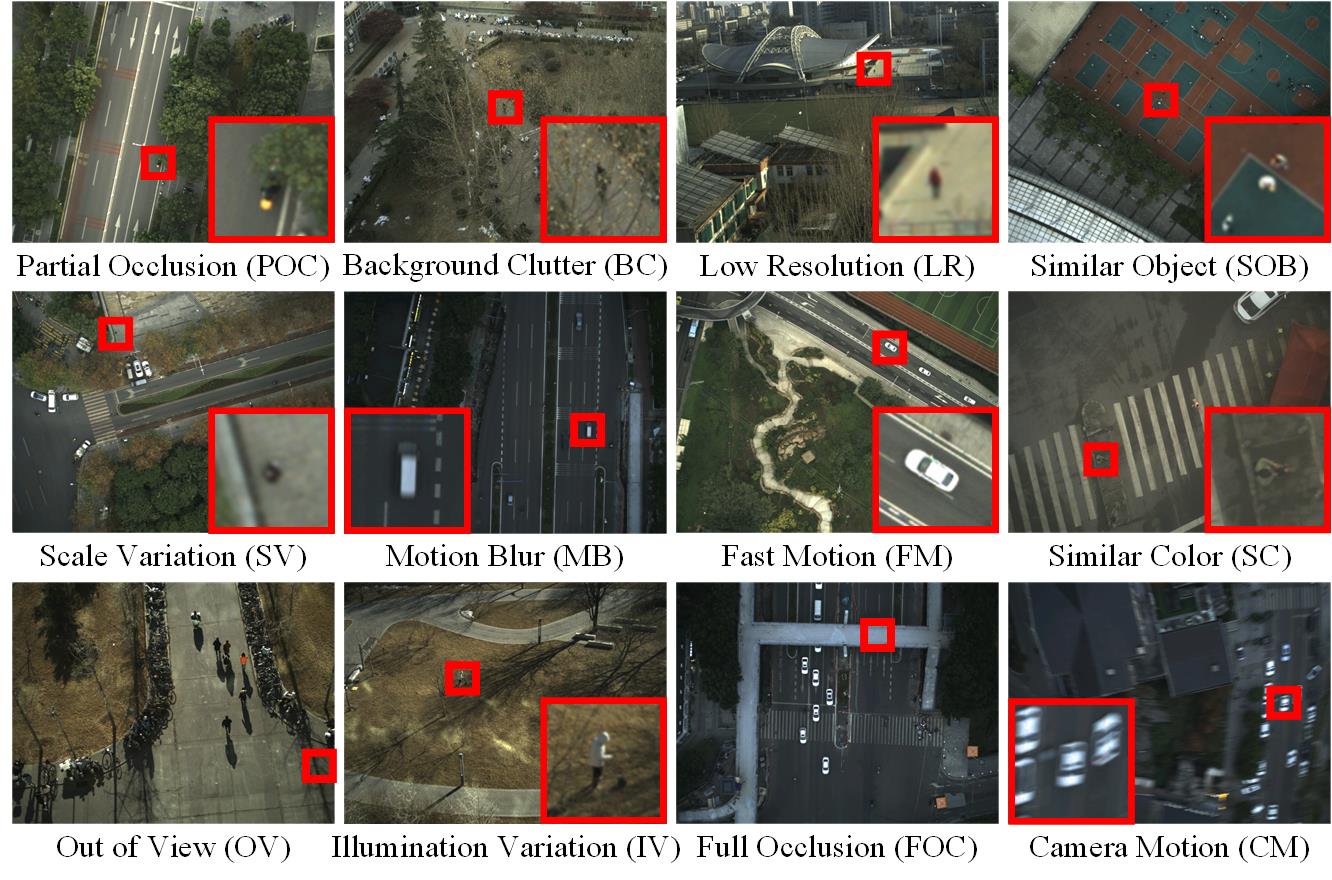}
  \caption{Examples of representative challenge scenarios, with targets marked by red boxes and displayed in magnified views.}
  \label{fig:challenge}
\end{figure}

\subsection{Dataset Analysis}

\noindent\textbf{Challenge Attributes.} To reflect the critical aspects of UAV tracking, we have defined 12 key challenge attributes as illustrated in \cref{fig:challenge}. These attributes offer a comprehensive overview of UAV tracking, covering target position, background characteristics, and UAV status. These challenges also highlight the necessity of using spectral information for UAV tracking. A more detailed explanation of these challenges is provided in the supplementary material.

\begin{figure}[t]
  \centering
  \includegraphics[width=1\linewidth]{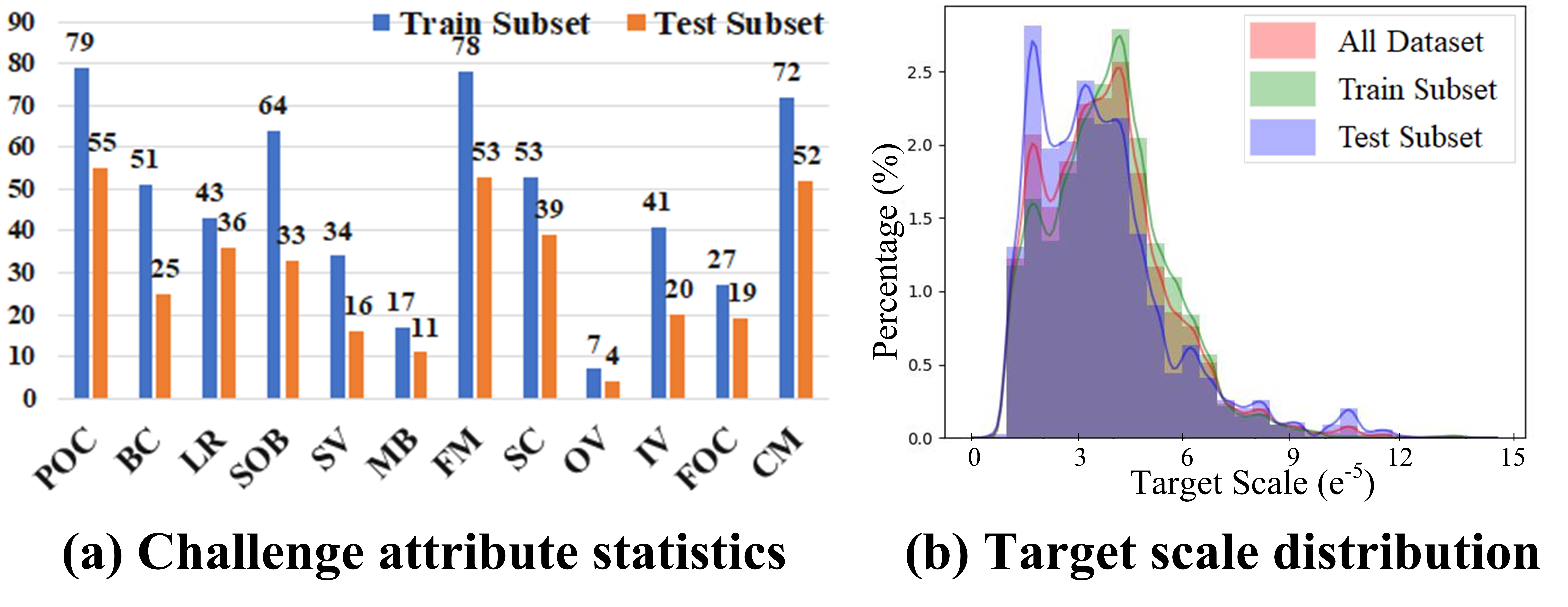}
  \caption{Statistics of the MUST dataset: (a) Distribution of video sequences across 12 key challenges, (b) Target scale distribution and consistency across different subsets.}
  \label{fig:statistics}
\end{figure}

\noindent \textbf{Statistical Analysis.}
We visualize the statistical properties of the MUST dataset to provide a comprehensive analysis. As shown in \cref{fig:statistics} (a), the 12 key challenge attributes are evenly distributed, ensuring the difficulty and diversity of both subsets. Moreover, we count the target scale distribution, as depicted in \cref{fig:statistics} (b). Most targets are small-sized, with their sizes accounting for around 3$\times10^{-5}$ of the entire image. The prevalence of small targets is due to the shooting altitude up to 200 meters, which brings new challenges.

\begin{figure*}[t]
  \centering
  \includegraphics[width=1.0\linewidth]{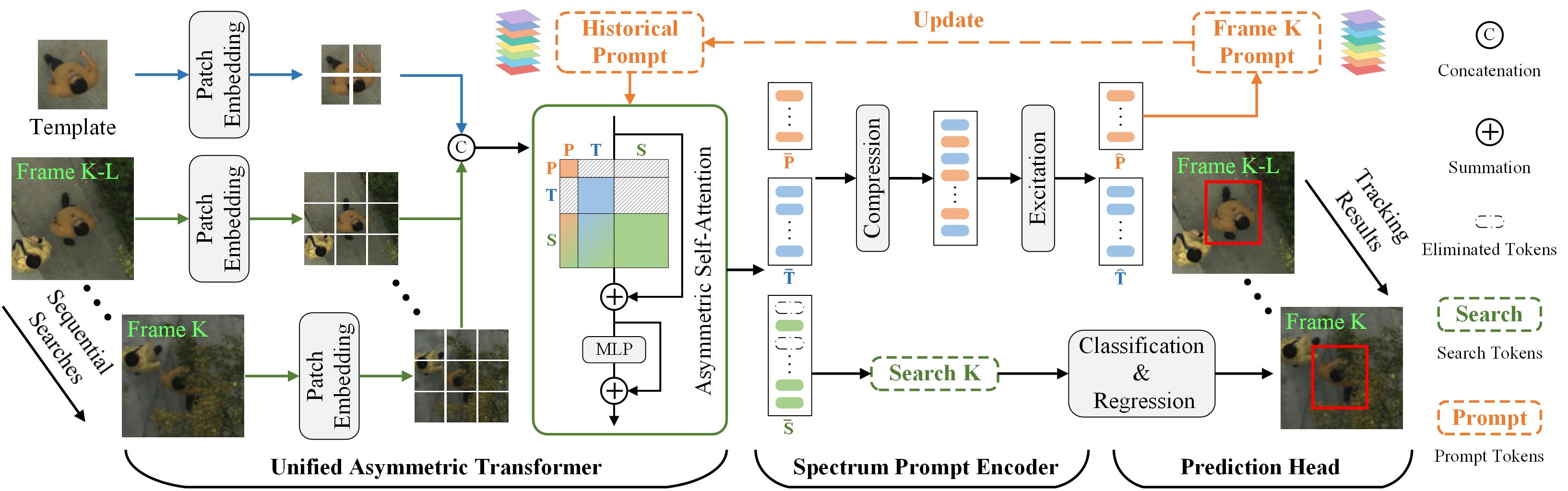}
  \caption{The proposed UNTrack consists of three components: the unified asymmetric transformer, the spectrum prompt encoder, and the prediction head. UNTrack takes the spectrum prompt, initial template, and sequential search as unified inputs and outputs the target bounding box for each frame. The encoded prompt tokens of the current frame update the spectrum prompt for subsequent tracking.}
  \label{fig:over}
\end{figure*}

\section{Method}

We propose the Unified Spectral-Spatial-Temporal Tracker (UNTrack) as a baseline for multispectral UAV tracking, as shown in \cref{fig:over}. UNTrack utilizes a unified asymmetric transformer to extract features from spectral, spatial, and temporal domains, while suppressing background noise in search regions. The pruned attention maps are fed into a spectrum prompt encoder, which generates prompts to deliver historical spectral characteristics. Finally, the prediction head updates the spectral-based prompt and outputs the target's position in the current frame.

\subsection{Unified Asymmetric Transformer}

\textbf{Revisit Transformer.} 
The vanilla transformer~\cite{vaswani2017attention} is commonly employed for feature extraction in one-stream trackers due to its exceptional long-range modeling ability. The input template and search are fed into a patch embedding layer and concatenated to form tokens $\textbf{\emph{F}}$. Subsequently, $\textbf{\emph{F}}$ is linearly projected to obtain query $\textbf{\emph{Q}}$, key $\textbf{\emph{K}}$, and value $\textbf{\emph{V}}$, with their respective linear projection weights $\textbf{\emph{W}}_{\rm{Q}}$, $\textbf{\emph{W}}_{\rm{K}}$, and $\textbf{\emph{W}}_{\rm{V}}$. The attention map $\textbf{\emph{A}}$ is computed as follows:
\begin{small}
\begin{align}
  & \textbf{\emph{Q}} = \textbf{\emph{F}}\textbf{\emph{W}}_{\rm{Q}}, \enspace \textbf{\emph{K}} = \textbf{\emph{F}}\textbf{\emph{W}}_{\rm{K}}, \enspace \textbf{\emph{V}} =    \textbf{\emph{F}}\textbf{\emph{W}}_{\rm{V}}, \\
  & \textbf{\emph{A}} = {\rm{Attn}}(\textbf{\emph{Q}},\textbf{\emph{K}},\textbf{\emph{V}}) = {\rm{Softmax}}(\frac{ \textbf{\emph{Q}} \textbf{\emph{K}}^{\top} }{\sqrt{d_{k}}}) \textbf{\emph{V}},
\end{align}
\end{small}
where $\sqrt{d_{k}}$ is a scaling factor. 

\noindent \textbf{Unified Embedding.}
The vanilla transformer focuses on spatial features but neglects spectral and temporal information. In contrast, the proposed UNTrack incorporates both sequential and historical input, enabling unified spectral-spatial-temporal feature extraction and relation modeling. Specifically, UNTrack initially treats consecutive frames as sequential searches $\textbf{\emph{I}}_{\rm{S}} = \{\textbf{\emph{I}}_{\rm{S_{\emph{i}}}}\in \mathbb{R}^{\rm{H_{S}} \times \rm{W_{S}} \times 8}|i=1,\dots,{\rm{N}}\}$ and embeds them together with the initial template $\textbf{\emph{I}}_{\rm{T}} \in \mathbb{R}^{\rm{H_{T}} \times \rm{W_{T}} \times 8}$. Here, $\textbf{\emph{I}}_{\rm{S_{\emph{i}}}}$ denotes each search frame, and $\rm{N}$ is the total frame number. The process is expressed as:
\begin{small}
\begin{align}
    &\textbf{\emph{T}} = \phi(\textbf{\emph{I}}_{\rm{T}}) \in \mathbb{R}^{\rm{L_{T}} \times \rm{C}}, \\
    &\textbf{\emph{S}} = [\phi(\textbf{\emph{I}}_{\rm{S_{1}}}); \dots; \phi(\textbf{\emph{I}}_{\rm{S_{N}}})] \in \mathbb{R}^{\rm{L_{S}} \times \rm{C}},
\end{align}
\end{small}
where $\textbf{\emph{T}}$ and $\rm{L_{T}}=\frac{H_T \times W_T}{16}$ denote the template tokens and their length. $\textbf{\emph{S}}$ and $\rm{L_{S}}=\frac{N \times H_S \times W_S}{16}$ represent the concatenated search tokens and their total length. $\rm{C}$ is the feature dimension of the tokens, and $\phi$ denotes the separation and flattening operations. Furthermore, UNTrack concatenates the historical prompt tokens $\textbf{\emph{P}} \in \mathbb{R}^{\rm{L_{P}} \times \rm{C}}$ from the previous frame, where $\rm{L_{P}}$ represents the token length, typically set to 1 in practice. The resulting $\textbf{\emph{F}} = [\textbf{\emph{P}};\textbf{\emph{T}};\textbf{\emph{S}}]$ is then used for subsequent attention calculations.

\begin{figure}[t]
  \centering
  \includegraphics[width=1\linewidth]{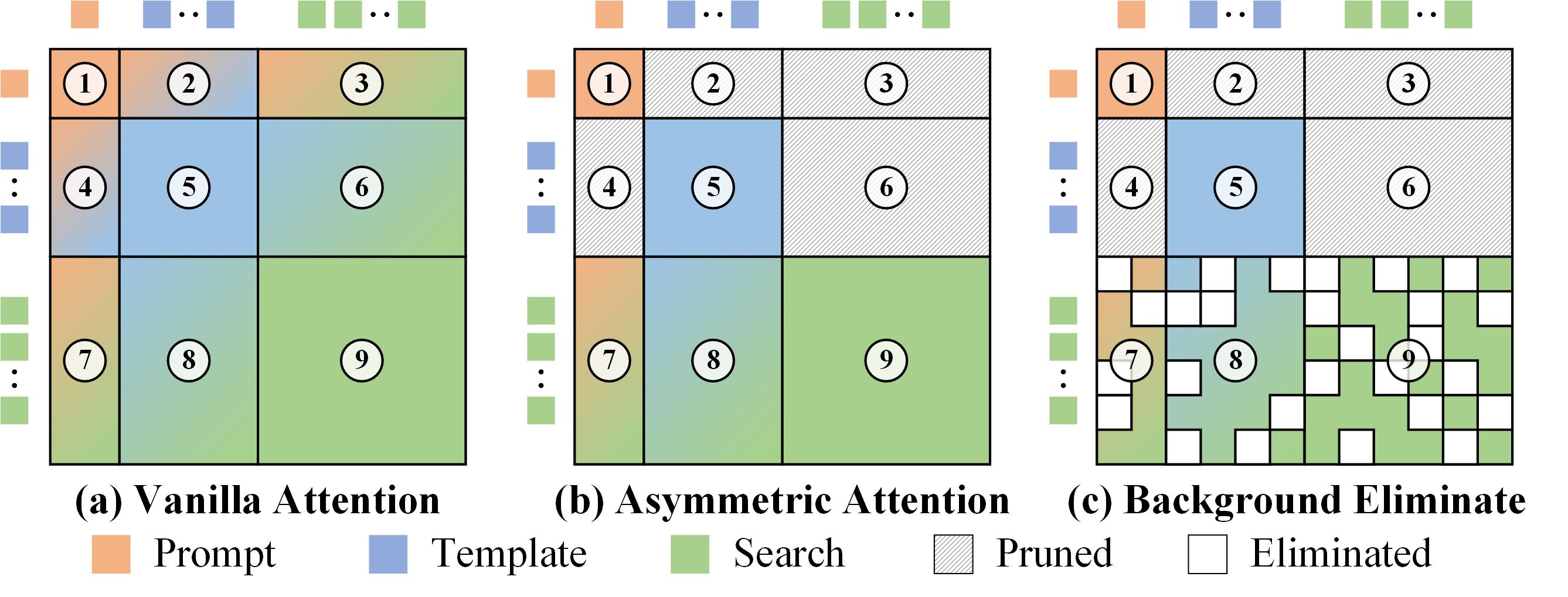}
  \caption{Information streams in the attention mechanism. The interaction between prompt, template, and search corresponds to the nine blocks in the attention map.}
  \label{fig:trans}
\end{figure}

\noindent \textbf{Asymmetric Attention.} 
Different from previous trackers, UNTrack adopts the unified tokens $\textbf{\emph{F}}$, resulting in a modified attention calculation formula, as follows:
\begin{small}
\begin{align}
    &\textbf{\emph{Q}} = [\textbf{\emph{Q}}_{\rm{P}};\textbf{\emph{Q}}_{\rm{T}};\textbf{\emph{Q}}_{\rm{S}}],
    \textbf{\emph{K}} = [\textbf{\emph{K}}_{\rm{P}};\textbf{\emph{K}}_{\rm{T}};\textbf{\emph{K}}_{\rm{S}}], \textbf{\emph{V}} = [\textbf{\emph{V}}_{\rm{P}};\textbf{\emph{V}}_{\rm{T}};\textbf{\emph{V}}_{\rm{S}}], \\
    &\textbf{\emph{A}} = {\rm{Softmax}}(\frac{[\textbf{\emph{Q}}_{\rm{P}}; \textbf{\emph{Q}}_{\rm{T}}; \textbf{\emph{Q}}_{\rm{S}}] [\textbf{\emph{K}}_{\rm{P}}; \textbf{\emph{K}}_{\rm{T}}; \textbf{\emph{K}}_{\rm{S}}]^{\top}}{\sqrt{d_{k}}}) [\textbf{\emph{V}}_{\rm{P}}; \textbf{\emph{V}}_{\rm{T}}; \textbf{\emph{V}}_{\rm{S}}].
\end{align}
\end{small}
According to the information streams pipeline, the calculated attention map $\textbf{\emph{A}}$ is divided into nine blocks. As depicted in \cref{fig:trans} (a), each block corresponds to the attention results of different token interactions, and their detailed descriptions are offered in the supplementary material.

Notably, each block is unevenly relevant to the tracking process. For instance, blocks 1, 5, and 9 denote the self-attention of tokens, extracting deep semantic features and representing target characteristics. Blocks 7 and 8 capture the cross-attention between the search and target, which is responsible for locating essential search regions. Conversely, the remaining blocks focus on the prompt and template, which are not only irrelevant to the tracking process but also brings computational cost and noise interference.

Inspired by the above insights, we propose an asymmetric attention mechanism, as shown in \cref{fig:trans} (b). Specifically, we prune the cross-attention on the prompt and template, and the attention map can be calculated as:
\begin{small}
\begin{align}
    \textbf{\emph{A}} =  [&{\rm{Attn}}(\textbf{\emph{Q}}_{\rm{P}},\textbf{\emph{K}}_{\rm{P}},\textbf{\emph{V}}_{\rm{P}}), {\rm{Attn}}(\textbf{\emph{Q}}_{\rm{T}},\textbf{\emph{K}}_{\rm{T}},\textbf{\emph{V}}_{\rm{T}}), \notag \\
    & {\rm{Attn}}(\textbf{\emph{Q}}_{\rm{s}},\textbf{\emph{K}},\textbf{\emph{V}})].
\end{align}
\end{small}
The proposed asymmetric structure preserves necessary attention maps, facilitating spectral-spatial-temporal features extraction at low cost. 

\noindent \textbf{Spectral Background Eliminate.} Based on asymmetric attention, UNTrack introduces a spectral background eliminate mechanism as shown in \cref{fig:trans} (c). The mechanism aims to progressively eliminate search regions belonging to the background, thereby reducing computational overhead and enhancing focus on the target. Specifically, the attention map $\textbf{\emph{A}}_{\rm{S}}\in \mathbb{R}^{({\rm{L_P}}+{\rm{L_T}}+{\rm{L_S}}) \times C}$ for the search tokens describes the correlation of each search region with the target from different perspectives. 

This provides a feasibility to quantify the confidence of each search region belonging to the target by calculating a scalar value, which is given by:
\begin{small}
\begin{align}
    B = \Vert \rm{AvgPool}(\textbf{\emph{A}}_{\rm{S}}) \Vert,
\end{align}
\end{small}
where $\rm{AvgPool}$ refers to the average pooling along the token length, and $\Vert \enspace \Vert$ denotes the Euclidean norm along the spectral dimension. A region is more likely to be background if its value is relatively small. Therefore, we keep the top $\rho \times ({\rm{L_P}}+{\rm{L_T}}+{\rm{L_S}})$ regions as candidates, while eliminating the remaining search regions. Here, $\rho$ indicates the keeping ratio, which is dynamically adjusted using a cosine annealing strategy during training to progressively eliminate background.

\subsection{Spectrum Prompt Encoder}
In multispectral UAV tracking, spectral information facilitates background discrimination and plays a pivotal role in addressing various challenges. To take advantage of this, UNTrack adopts a spectral prompt encoder, as illustrated in \cref{fig:over}. The encoder takes the unified tokens $\bar{\textbf{\emph{F}}}=[\bar{\textbf{\emph{P}}};\bar{\textbf{\emph{T}}};\bar{\textbf{\emph{S}}}]$ as input, which are obtained from the asymmetric transformer. Here, $\bar{\textbf{\emph{P}}}$, $\bar{\textbf{\emph{T}}}$, and $\bar{\textbf{\emph{S}}}$ correspond to the generated prompt, template, and search tokens, respectively. The prompt and template tokens are then concatenated to generate $[\bar{\textbf{\emph{P}}}; {\rm{GPool}}(\bar{\textbf{\emph{T}}})] \in \mathbb{R}^{{\rm{L_{P}}} \times 2C}$ where $\rm{GPool}$ denotes global pooling along the token length, which highlights the target's spectral features.

Subsequently, UNTrack employs CNN-based layers to process the spectral features of the prompt and template, enabling attention interaction. This process is described as:
\begin{small}
\begin{align}
    & \textbf{\emph{A}}' = {\rm{FC}}([\bar{\textbf{\emph{P}}}; {\rm{GPool}}(\bar{\textbf{\emph{T}}})]) \in \mathbb{R}^{{\rm{L_{P}}} \times \frac{{\rm{2C}}}{r}}, \\
    & \textbf{\emph{A}}'' = {\rm{FC}}(\textbf{\emph{A}}') \in \mathbb{R}^{{\rm{L_{P}}} \times {\rm{2C}}}, \\
    & [\widehat{\textbf{\emph{P}}};\widehat{\textbf{\emph{T}}}] = \textbf{\emph{A}}'' + {\rm{MLP}}(\textbf{\emph{A}}'') \in \mathbb{R}^{{\rm{L_{P}}} \times {\rm{2C}}},
\end{align}
\end{small}
where $r$ denotes the compression ratio, and both $\rm{FC}$ layers are implemented via convolution. The first $\rm{FC}$ layer performs spectral compression, while the second $\rm{FC}$ layer is responsible for excitation. The compression-excitation operation enables spectral attention calculation for the prompt and template. The $\rm{MLP}$ denotes a multi-layer perceptron, which does not alter the spectral dimensions. The final encoded $\widehat{\textbf{\emph{P}}} \in \mathbb{R}^{{\rm{L_{P}}} \times {\rm{C}}}$, recording the target's material properties, is regarded as the spectrum prompt for subsequent frame tracking.

\subsection{Prediction Head and Loss Function}
UNTrack adopts a dual-branch structure prediction head, which processes the search tokens $\bar{\textbf{\emph{S}}}$ as input. The head initially re-interprets tokens into 2D spatial feature maps and then feeds them into two separate branches responsible for classification and regression. The classification branch outputs a confidence map ${\rm{Cla}} \in \mathbb{R}^{1 \times {\rm{H_{S}}} \times {\rm{W_{S}}}}$, while the regression branch predicts a bounding box ${\rm{BBox}} \in \mathbb{R}^{4 \times {\rm{H_{S}}} \times {\rm{W_{S}}}}$, both generated through stacked convolutional layers.

During training, we adopt focal loss~\cite{law2018cornernet} for classification, denoted as $\mathcal{L}_{\rm{cls}}$. For bounding box regression, we employ both $\mathcal{L}_1$ loss and $\rm{GIoU}$ loss~\cite{rezatofighi2019generalized} (denoted as $\mathcal{L}_{\rm{GIoU}}$). The total loss $\mathcal{L}$ for model convergence is defined as:
\begin{small}
\begin{align}
    \mathcal{L} = \mathcal{L}_{\rm{cls}} + \lambda_{1}\mathcal{L}_{1} + \lambda_{2}\mathcal{L}_{\rm{GIoU}},
\end{align}
\end{small}
where $\lambda_{1}=5$ and $\lambda_{2}=2$ are the weighting factors that balance the contribution of each loss term.

\section{Experiment}

\subsection{Implementation Details}
\label{sec:details}
We train and evaluate numerous state-of-the-art trackers on the proposed Multispectral UAV Single Object Tracking (MUST) dataset, each with fine-tuned hyperparameters for optimal performance. Moreover, we evaluate our Unified Spectral-Spatial-Temporal Tracker (UNTrack) across multiple datasets. We adopt the AdamW optimizer with an initial learning rate of $10^{-4}$, resizing searches to $384 \times 384$ and templates to $192 \times 192$. We train for 50 epochs with a batch size of 24 and reduce the learning rate by a factor of 10 after 30 epochs. All experiments are conducted on NVIDIA RTX3090 GPUs.

Due to differences in spectral resolution and range across datasets, there are no available MSI-based pre-trained parameters. We thus load parameters from the ImageNet dataset~\cite{deng2009imagenet} and adopt a simple yet effective parameter reconstruction strategy. The strategy aims to extend RGB-based pre-trained parameters to MSI-based vision tasks through interpolation. We provide a detailed description and analysis of this strategy in the supplementary material. \Cref{tab:methods} presents the performance of trackers with different parameter sources, with * marking the reconstructed parameters and otherwise loading RGB-based parameters directly.

\subsection{Results on the MUST Dataset}

\textbf{Main Results.} We compare UNTrack against other state-of-the-art trackers on the MUST dataset, as shown in \cref{tab:methods}. Obviously, one-stream trackers outperform Siamese trackers and have become the dominant approach in visual tracking tasks. Additionally, the parameter reconstruction strategy significantly improves the accuracy, with an average increase of over $10\%$ across metrics, as confirmed on various trackers. Compared with other trackers, UNTrack achieves state-of-the-art performance, for example, surpassing OSTrack~\cite{ye2022joint} by $4.6\%$ in AUC and $5.9\%$ in Precision (Pre).

\begin{table}
  \caption{Comparison with state-of-the-art trackers on the MUST dataset. * denotes trackers trained with reconstructed parameters. The top two results are highlighted in \textcolor{red}{red} and \textcolor{blue}{blue}.}
  \label{tab:methods}
  \small
  \centering
  \setlength{\tabcolsep}{6pt}
  \begin{tabular}{l|ccccc}
    \toprule
    Method & AUC & SR$_{0.5}$ & SR$_{0.75}$ & Pre & Pre$_{N}$ \\
    \midrule
    SiamRPN~\cite{li2018high} & 38.9 & 49.5 & 22.3 & 56.7 & 50.1 \\
    SiamMask~\cite{wang2019fast} & 39.1 & 51.0 & 25.3 & 56.1 & 52.8 \\
    DiMP~\cite{bhat2019learning} & 47.3 & 60.7 & 31.6 & 69.8 & 57.1 \\
    TaMOs~\cite{mayer2024beyond} & 20.4 & 24.5 & 12.0 & 31.0 & 23.8 \\
    ToMP~\cite{mayer2022transforming} & 30.3 & 39.1 & 14.5 & 46.1 & 40.5 \\
    SGDViT~\cite{yao2023sgdvit} & 38.1 & 50.2 & 22.8 & 54.4 & 52.1 \\
    OSTrack$_{256}$~\cite{ye2022joint} & 39.9 & 50.5 & 27.4 & 55.9 & 51.2 \\
    OSTrack$_{384}$~\cite{ye2022joint} & 44.5 & 56.7 & 27.1 & 63.9 & 56.8 \\
    ODTrack~\cite{zheng2024odtrack} & 46.3 & 58.4 & 28.1 & 67.6 & 60.4 \\
    ZoomTrack~\cite{kou2024zoomtrack} & 44.0 & 56.2 & 25.2 & 63.5 & 55.9 \\
    \midrule
    OSTrack$_{256}^{*}$~\cite{ye2022joint} & 52.3 & 64.8 & 36.3 & 67.5 & 65.2 \\
    OSTrack$_{384}^{*}$~\cite{ye2022joint} & \textcolor{blue}{55.1} & \textcolor{blue}{69.6} & \textcolor{blue}{44.1} & \textcolor{blue}{73.3} & \textcolor{blue}{68.8} \\
    ODTrack$^{*}$~\cite{zheng2024odtrack} & 52.7 & 67.3 & 38.8 & 71.3 & 66.2 \\
    ZoomTrack$^{*}$~\cite{kou2024zoomtrack} & 53.3 & 67.7 & 40.4 & 72.3 & 66.7 \\
    \midrule  
    \textbf{UNTrack (Ours)} & 47.8 & 61.0 & 28.9 & 68.3 & 59.4 \\
    \textbf{UNTrack$^{*}$ (Ours)} & \textcolor{red}{59.7} & \textcolor{red}{75.8} & \textcolor{red}{46.2} & \textcolor{red}{79.2} & \textcolor{red}{74.8}\\
    \bottomrule
  \end{tabular}
\end{table}

\begin{figure}[t]
  \centering
  \includegraphics[width=1.0\linewidth]{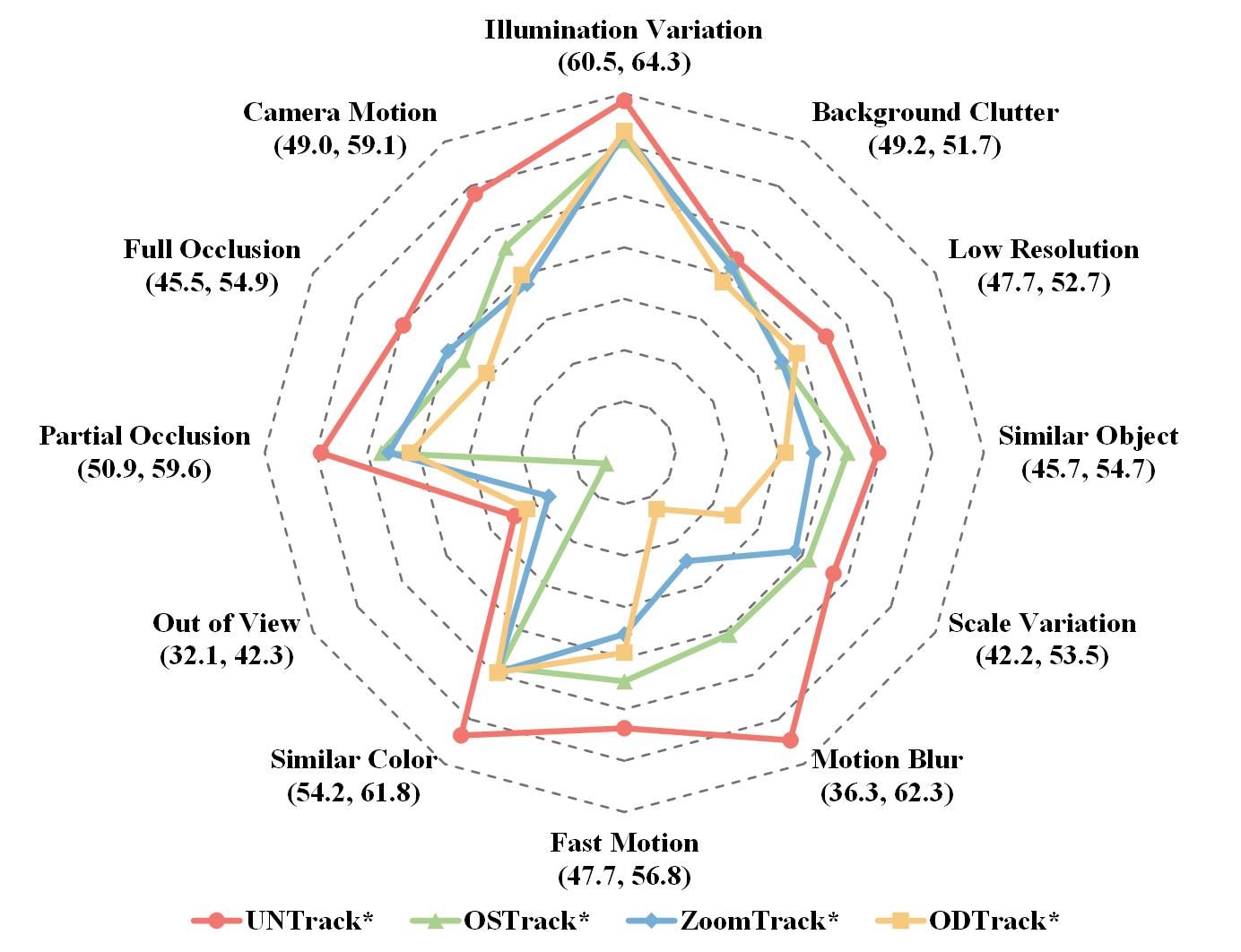}
  \caption{A comparison of UNTrack with other state-of-the-art trackers in terms of AUC across each challenge attribute on MUST dataset, with the lowest and highest performance values marked.}
  \label{fig:leida}
\end{figure}

\noindent\textbf{Attribute-based Comparison.}
Furthermore, we compare the performance of various trackers on each challenge attribute, as illustrated in \cref{fig:leida}. Compared to other trackers, UNTrack achieves superior accuracy across all challenges. Notably, UNTrack excels under demanding motion conditions such as MB and CM. UNTrack's robustness is primarily due to its emphasis on the spectrum, which remains undisturbed by violent motion, thereby reducing tracking drift. Additionally, UNTrack leverages unified spectral-spatial-temporal features, demonstrating exceptional potential in handling challenges with restricted appearance, such as SC and FOC. When handling the most challenging OV attribute, UNTrack employs historical prompt and sequential searches enabling successful re-tracking despite short-term target loss, and outperforming previous trackers.

Evaluation of state-of-the-art trackers on the MUST dataset demonstrates its feasibility as reliable data support. Notably, the defined 12 key challenge attributes emphasize the unique characteristics of multispectral UAV data, laying the groundwork for targeted improvements in trackers. Moreover, extensive experiments show the superior of UNTrack. In conclusion, we have established a pioneering benchmark for multispectral UAV tracking. 

\begin{figure}[t]
  \centering
  \includegraphics[width=1\linewidth]{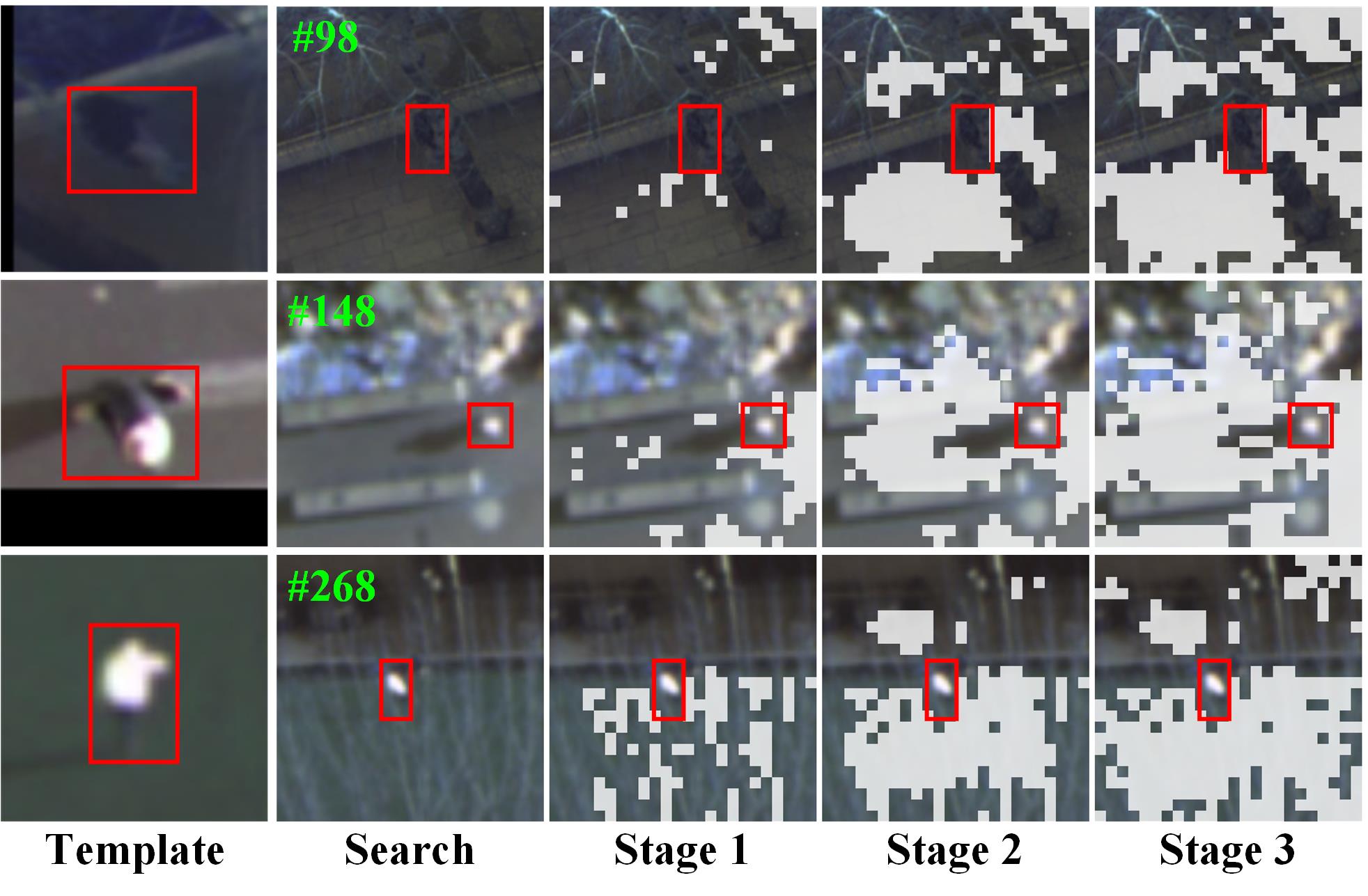}
  \caption{Visualization of the progressive background elimination process using the spectral background elimination mechanism. The target is highlighted with a red box, and the white masks indicate the eliminated regions.}
  \label{fig:sc}
\end{figure}

\noindent\textbf{Visualization of Background Elimination.}
To intuitively demonstrate the superiority of UNTrack, we visualize its behavior for qualitative analysis. We first present the background elimination process at different stages in the asymmetric transformer, providing an in-depth analysis of the spectral background eliminate mechanism, as shown in \cref{fig:sc}. UNTrack gradually eliminates search regions belonging to the background based on spectral confidence, greatly reducing computational overhead.

\begin{figure}[t]
  \centering
  \includegraphics[width=1\linewidth]{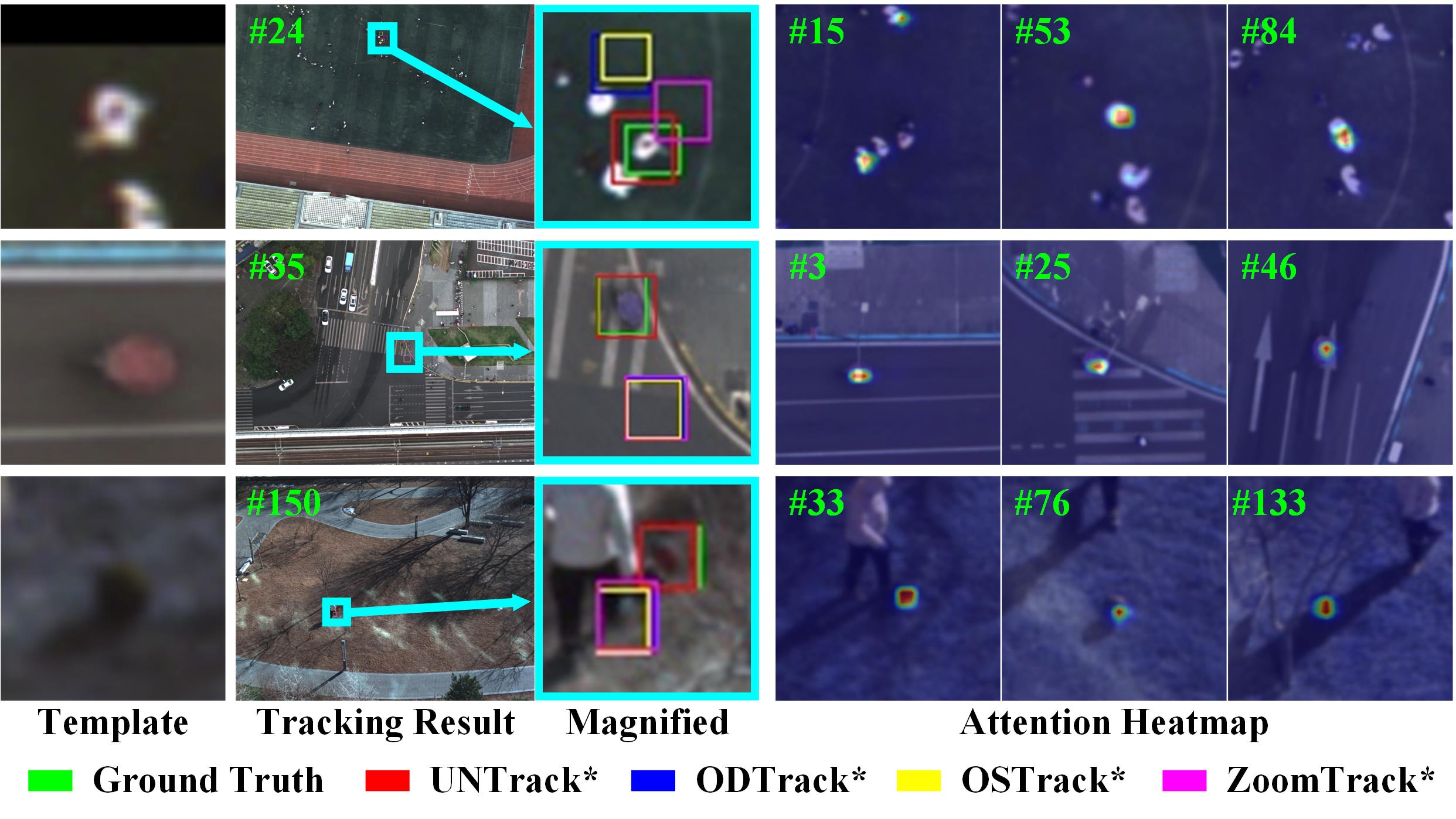}
  \caption{Comparison of tracking results for UNTrack and other state-of-the-art trackers, with the target area magnified. The attention heatmap highlights the regions of focus for UNTrack.}
  \label{fig:attn}
\end{figure}

\noindent\textbf{Visualization of Tracking Results.}
Furthermore, we compare the tracking performance of UNTrack against various trackers, as shown in \cref{fig:attn}. Despite the presence of objects with similar appearances or backgrounds of similar colors around the target, UNTrack consistently delivers robust tracking results, surpassing other trackers. In addition, attention heatmaps illustrate the focus regions of UNTrack in each search frame. Obviously, UNTrack maintains a steadfast focus on the target amidst various challenges, an ability enabled by leveraging spectral features.

\begin{figure}[t]
  \centering
  \includegraphics[width=1\linewidth]{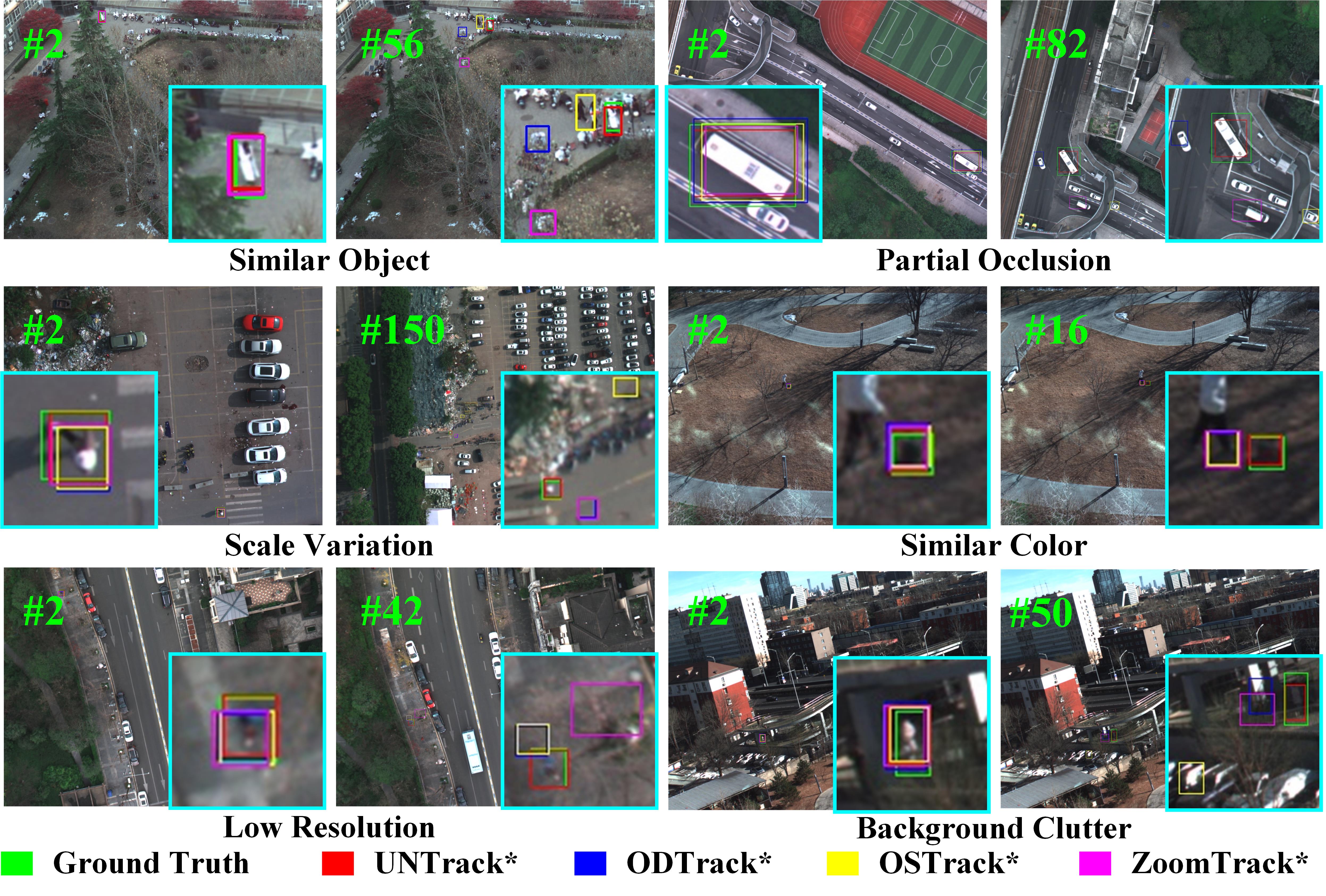}
  \caption{Comparison of UNTrack and other state-of-the-art trackers when dealing with various challenge attributes.}
  \label{fig:cha}
\end{figure}

We visualize the tracking results for various challenge attributes that are intractable for traditional RGB-based trackers. As shown in \cref{fig:cha}, RGB-based trackers struggle to distinguish the target when it is surrounded by similar objects or background. When the target is occluded, the template features extracted by RGB-based trackers are contaminated, leading to tracking drift. In addition, RGB-based trackers cannot obtain sufficient spatial features from the small-sized target, thus performing poorly. In contrast, UNTrack adopts the novel spectral-spatial-temporal unified tracking framework, providing stable tracking results across all challenges.

\subsection{Versatility Evaluation}

\noindent \textbf{MSI-Based General Tracking.}
The HOT~\cite{xiong2020material} is an available dataset used for MSI-based general tracking. We conduct extensive comparative experiments based on it and additionally introduce Frames Per Second (FPS) as a metric for tracker efficiency. For a fair comparison, all trackers perform inference on single NVIDIA RTX 3090 GPU. As shown in \cref{tab:hot}, UNTrack achieves the fastest inference speed at 37 FPS while also reaching state-of-the-art tracking accuracy. This optimal trade-off between accuracy and efficiency is enabled by the asymmetric attention structure and spectral background eliminate mechanism, which effectively eliminate redundant tokens.

\begin{table}
  \caption{Comparison of results for UNTrack and other state-of-the-art trackers on the HOT dataset, with the top two results highlighted in \textcolor{red}{red} and \textcolor{blue}{blue}.}
  \label{tab:hot}
  \small
  \centering
  \setlength{\tabcolsep}{15pt}
  \begin{tabular}{l|ccc}
    \toprule
    Method & AUC & Pre & FPS \\
    \midrule
    MHT~\cite{xiong2020material} & 58.7 & 88.0 & \textendash  \\ 
    BAENet~\cite{li2020bae} & 60.6 & 87.7 & \textendash  \\ 
    SSTNet~\cite{li2021spectral} & 62.1 & 90.1 & \textendash  \\ 
    SiamHYPER~\cite{liu2022siamhyper} & 67.8 & \textcolor{blue}{94.5} & \textcolor{blue}{27.7}  \\
    TMTNet~\cite{zhao2023tmtnet} & \textcolor{blue}{69.9} & 92.8 & 12.6  \\
    SSTFT~\cite{wang2023spectral} & 68.2 & 88.4 & \textendash  \\ 
    SEENet~\cite{li2023learning} & 65.3 & 93.0 & 13.1  \\
    HANet~\cite{liu2024deep} & 68.8 & \textcolor{red}{94.8} & 21.2  \\
    \midrule
    \textbf{UNTrack (Ours)} & \textcolor{red}{70.4} & 93.7 & \textcolor{red}{37.0} \\
    \bottomrule
  \end{tabular}
\end{table}

\noindent \textbf{RGB-Based General Tracking.} 
We extend UNTrack to the GOT10K dataset~\cite{huang2019got}, which is commonly used for RGB-based general tracking, providing a more comprehensive analysis. As shown in \cref{tab:got}, UNTrack achieves performance comparable to other state-of-the-art trackers, slightly falling short on partial metrics. This is primarily due to the limited spectral information provided by RGB images, which restricts the performance of the spectrum prompt encoder. However, the asymmetric attention structure still enables robust feature extraction, demonstrating UNTrack's potential across different data modalities.

\subsection{Ablation Study}
To perform a thorough analysis of the innovations introduced by UNTrack, we have conducted rigorous ablation studies on the proposed MUST dataset. All experiments adopt uniform training settings, as detailed in \cref{sec:details}, to ensure consistency and isolate variables.

\begin{table}
  \caption{The performance of our UNTrack and other state-of-the-art trackers on the GOT10K dataset. The best two results are highlighted in \textcolor{red}{red} and \textcolor{blue}{blue}.}
  \label{tab:got}
  \small
  \centering
  \setlength{\tabcolsep}{14pt}
  \begin{tabular}{l|ccc}
    \toprule
    Method  & AO & SR$_{0.5}$ & SR$_{0.75}$ \\
    \midrule
    SiamRPN++~\cite{li2019siamrpn++}  & 51.7 & 61.6 & 32.5 \\
    AutoMatch~\cite{zhang2021learn}  & 65.2 & 76.6 & 54.3 \\
    OSTrack~\cite{ye2022joint}  & 71.0 & 80.4 & 68.2 \\
    MixFormer~\cite{cui2022mixformer}  & 72.5 & 82.4 & 69.9 \\
    CTTrack~\cite{song2023compact}  & 71.3 & 80.7 & 70.3 \\
    SeqTrack~\cite{chen2023seqtrack} &  74.7 & 84.7 & 71.8 \\
    LoRAT~\cite{lin2025tracking}  & 72.1 & 81.8 & 70.7 \\
    ARTrackV2~\cite{bai2024artrackv2} &  75.9 & 85.4 & 72.7 \\
    ODTrack~\cite{zheng2024odtrack}  & 77.0 & 87.9 & \textcolor{red}{75.1} \\
    HIPTrack~\cite{cai2024hiptrack}  & \textcolor{red}{77.4} & \textcolor{blue}{88.0} & 74.5 \\
    \midrule
    \textbf{UNTrack (Ours)} & \textcolor{blue}{77.3} & \textcolor{red}{88.4} & \textcolor{blue}{74.7} \\
    \bottomrule
  \end{tabular}
\end{table}

\begin{table}
  \caption{Ablation study on the number of sequential search frames.}
  \label{tab:length}
  \small
  \centering
  \setlength{\tabcolsep}{9pt}
  \begin{tabular}{c|ccccc}
    \toprule
    Number & AUC & SR$_{0.5}$ & Pre & FLOPs & FPS \\
    \midrule
    1 & 54.6 & 70.2 & 73.6 & 74.1G & 54.6 \\
    2 & 59.7 & 75.8 & 79.2 & 106.9G & 38.0 \\
    3 & 59.9 & 75.8 & 79.3 & 139.7G & 20.5 \\
    4 & 58.5 & 73.9 & 76.1 & 172.5G & 12.6 \\
    5 & 56.3 & 71.2 & 72.8 & 205.3G & 9.2 \\
    \bottomrule
  \end{tabular}
\end{table}

\noindent \textbf{Number of Search Frames.} 
UNTrack processes sequential searches instead of conventional single-frame search, thereby enabling unified feature extraction. As depicted in \cref{tab:length}, we evaluate the influence of varying search frame numbers on tracking performance. When transitioning from single-frame to sequence-based input, the tracker experiences a notable increase in AUC by $5.1\%$. However, further frame increments do not consistently translate into performance gains, and the computational burden is escalating. UAV tracking often involves drastic changes in the target state between consecutive frames, making excessive search frames likely to introduce learning burdens and potentially noise. Therefore, we made a trade-off between accuracy and efficiency, setting the number of search frames to 2.

\noindent \textbf{Asymmetric Structure Configuration.}
To analyze the proposed asymmetric transformer, we compare different structural configurations, as shown in \cref{tab:asymmetric}. Obviously, the introduction of spectrum prompt effectively improves AUC by $1.6\%$, while slowing down inference speed. We thus further introduce an asymmetric attention structure to prune cross-attention for both the prompt and template. The structural innovation greatly improves AUC by $4.1\%$ while reducing FLOPs by $19.0\%$. Moreover, we compare the tracking performance before and after enabling the spectral background eliminate mechanism, which improves inference efficiency by progressively eliminating background while preserving tracking accuracy.

\begin{table}
  \caption{Comparison of different attention structures. SA and CA refer to self-attention and cross-attention applied to the prompt (P), template (T), and search (S). SC represents the introduced spectral background elimination mechanism.}
  \label{tab:asymmetric}
  \small
  \centering
  \setlength{\tabcolsep}{6pt}
  \begin{tabular}{ccc|ccc|c|ccc}
    \toprule
    \multicolumn{3}{c|}{SA} & \multicolumn{3}{c|}{CA} & \multirow{2}{*}{SC} & \multirow{2}{*}{AUC} & \multirow{2}{*}{FLOPs} & \multirow{2}{*}{FPS} \\
     \cline{1-6}
    \multicolumn{1}{c}{P} & \multicolumn{1}{c}{T} & \multicolumn{1}{c|}{S} & \multicolumn{1}{c}{P} & \multicolumn{1}{c}{T} & \multicolumn{1}{c|}{S} & & & & \\
    \midrule
    \textendash & \checkmark & \checkmark & \textendash & \checkmark & \checkmark & \textendash & 53.8 & 140.9G & 23.9 \\
    \checkmark & \checkmark & \checkmark & \checkmark & \checkmark & \checkmark & \textendash & 55.4 & 169.8G & 21.4 \\
    \checkmark & \checkmark & \checkmark & \textendash & \textendash & \checkmark & \textendash & 59.5 & 137.5G & 24.3 \\
    \checkmark & \checkmark & \checkmark & \textendash & \textendash & \checkmark & \checkmark & \textbf{59.7} & \textbf{106.9G} & \textbf{38.0} \\
    \bottomrule
  \end{tabular}
\end{table}

\noindent \textbf{Spectrum Prompt Source.}
UNTrack employs a spectrum prompt encoder responsible for recording and updating the target spectral characteristics. We compare different approaches of generating spectrum prompt, as depicted in \cref{tab:prompt}. In the absence of a spectrum prompt, UNTrack maintains a compact model size but suffers from suboptimal tracking accuracy. The prompt from random initialization without updating not only fails to enhance performance but also introduces side effects by introducing irrelevant information. Notably, leveraging historical prompt for updates markedly boosts performance, and the encoded spectrum prompts exhibit stronger target representation capabilities.

\begin{table}
  \caption{Comparison of spectrum prompt from different sources.}
  \label{tab:prompt}
  \centering
  \setlength{\tabcolsep}{6pt}
  \begin{tabular}{l|ccc}
    \toprule
    Source & AUC & \#Params & FPS \\
    \midrule
    None & 53.9 & 93.1M & 39.9 \\
    Random Initialization & 53.7 & 101.3M & 39.2 \\
    Asymmetric Transformer & 55.0 & 101.3M & 39.2 \\
    Spectrum Prompt Encoder & 59.7 & 112.6M & 38.0 \\
    \bottomrule
  \end{tabular}
\end{table}

\noindent \textbf{Spectral Encoder.} The proposed spectral encoder aims to generate prompts for subsequent tracking, which utilizes historical spectral features to improve performance. As shown \cref{tab:encoder}, we first implement a Naive encoder, which improves AUC by $3.7\%$ using only one convolution layer, proving the effectiveness of introducing a spectral encoder into network design. Notably, the Naive encoder struggles to generate robust prompts without spectral channels interaction. Therefore, we propose the Spectrum Prompt Encoder, which performs a compression-excitation operation on the spectral channels through FC and MLP layers. Our proposed encoder structure fully exploits spectral features and further improves AUC by $2.1\%$.

\begin{table}[h]
  \label{tab:encoder}
  \centering
  \setlength{\tabcolsep}{8pt}
  \begin{tabular}{c|ccccc}
    \toprule
    Encoder & AUC & SR$_{0.5}$ & SR$_{0.75}$ & Pre & Pre$_{N}$ \\
    \midrule
    w/o & 53.9 & 68.1 & 41.5 & 71.9 & 67.7 \\
    Naive & 57.6 & 72.4 & 45.3 & 76.6 & 70.8 \\
    Ours & 59.7 & 75.8 & 46.2 & 79.2 &74.8 \\
    \bottomrule
  \end{tabular}
\end{table}
\section{Conclusion}

This paper presented the first large-scale dataset for Multispectral UAV Single Object Tracking (MUST) and introduced UNTrack, a novel tracking framework that integrates spectral, spatial, and temporal features. UNTrack employs an asymmetric transformer with a spectral background eliminate mechanism and continuous spectrum prompt updates, achieving significant improvements in both tracking accuracy and efficiency. We believe the MUST dataset and the UNTrack framework will catalyze future advancements in UAV tracking research.

\noindent\textbf{Limitations and Future Work} Due to the principles and cost limitations of multispectral imaging, obtaining high spectral resolution video sequences at high frame rates is not trivial. In future work, we will collect video sequences with additional spectral bands and higher frame rates to address more challenging scenarios.


\clearpage
\setcounter{page}{1}
\maketitlesupplementary

\section{Dataset Introduction}

\subsection{Collection}
\begin{figure}[t]
  \centering
  \includegraphics[width=0.9\linewidth]{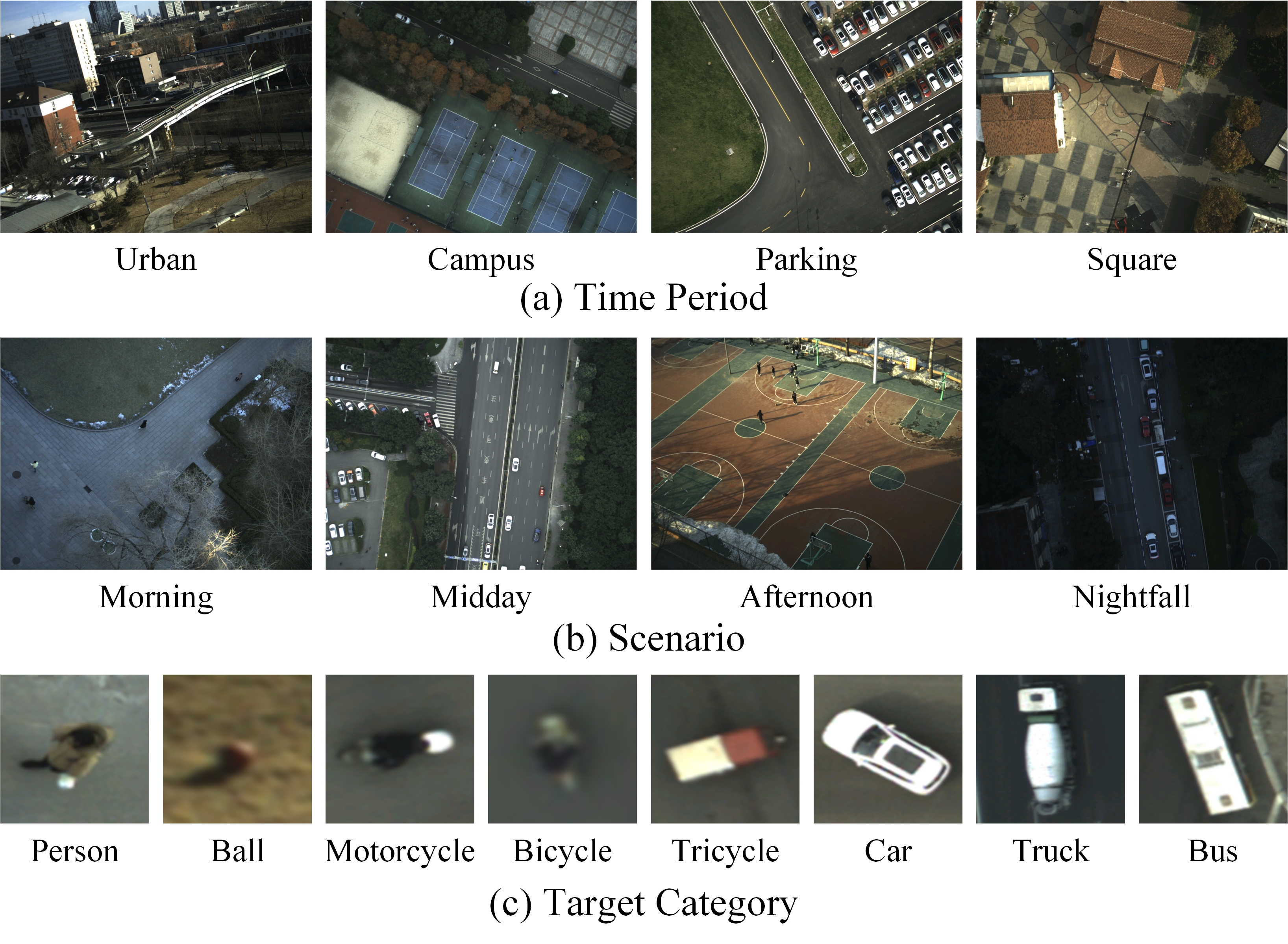}
  \caption{Eight target categories collected across various scenarios and time periods.}
  \label{fig:scences}
\end{figure}

The rapid advancements in UAV and sensor technologies have enabled the collection of multispectral video sequences from UAV platforms. In this study, we employed a multispectral camera, which captures 8 spectral bands in the wavelength range of 390–950 nm, as illustrated in \cref{tab:equipment}. Additionally, the camera is capable of recording video at 5 frames per second, with a spatial resolution of $1280\times960$ pixels per frame. To ensure a diverse dataset, we performed data collection across various scenarios and time periods, as depicted in \cref{fig:scences} (a) and (b). Moreover, the UAV captured multi-angle views of several categories of common targets of interest in different flight postures, as shown in \cref{fig:scences} (c). During the data collection process, the UAV maintained an altitude ranging from 20 to 250 meters above ground level, allowing for targets of varying sizes.

\subsection{Processing}
The multispectral camera adopts array distributed sensors to capture data across different spectral bands. To ensure accurate alignment of the multi-band data, we first perform geometric correction on each individual channel, mitigating parallax errors caused by variations in sensor positioning. Given that geometric correction can alter the resolution of each channel, we crop and retain the central $1200\times900$ pixels region of each frame. Additionally, the substantial distance between the UAV and the ground target exacerbates the effects of solar radiation and atmospheric scattering on sensors. To address these issues, we apply radiometric correction to calibrate the sensor responses and obtain accurate spectral curves. After these corrections, each multispectral image frame is represented as $\textbf{\emph{I}} \in \mathbb{R}^{1200 \times 900 \times 8}$.

\begin{table}
  \caption{The distribution of spectral bands collected by the used multispectral camera.}
  \label{tab:equipment}
  \centering
  \setlength{\tabcolsep}{4pt}
  \begin{tabular}{c|cccc}
    \toprule
    Bands & Start (nm) & End (nm) & Center (nm) & Width (nm) \\
    1 & 395 & 450 & 422.5 & 55 \\
    2 & 455 & 520 & 487.5 & 65 \\
    3 & 525 & 575 & 550 & 50 \\
    4 & 580 & 625 & 602.5 & 45 \\
    5 & 630 & 690 & 660 & 60 \\
    6 & 705 & 745 & 725 & 40 \\
    7 & 750 & 820 & 785 & 70 \\
    8 & 825 & 950 & 887.5 & 125 \\
    \bottomrule
  \end{tabular}
\end{table}

We engaged domain experts to to annotate the dataset using the DarkLabel 2.3 toolbox. For each frame, we provided the target's bounding box and status flags as ground truth. The bounding box is represented as $[x_{min}, y_{min}, w, h]$, where $(x_{min}, y_{min})$ denote the coordinates of the upper-left corner, and $(w, h)$ indicates target shape. If the target is fully occluded or temporarily out of view, the status flags are set to 1, with the bounding box marked as $[0, 0, 0, 0]$. After thorough screening, we retained 250 video sequences, totaling 42,671 frames and approximately 8,534 seconds. Detailed dataset information is provided in Table \ref{dataset}.

We divide the train and test subsets primarily based on the principle that the data across subsets is unbiased and consistently distributed. Initially, we randomly divide the data into two subsets. Then we adjust them to ensure consistency in terms of average frames across sequences, target size distribution, and proportion of challenge attributes.

\begin{table*}
  \caption{Properties of the Multispectral UAV Single Object Tracking (MUST) Dataset.}
  \label{dataset}
  \centering
  \setlength{\tabcolsep}{4pt}
  \begin{tabular}{c|cc|cccc|ccc}
    \toprule
    \multirow{2}{*}{Name} & \multirow{2}{*}{Video} & \multirow{2}{*}{Duration} & \multicolumn{4}{c|}{Frames} & \multicolumn{3}{c}{Resolution}\\
     &  &  & Total & Minimal & Maximal & Mean & Spatial Pixels & Spectral Range & Band Numbers \\
    \midrule
    MUST & 250 & 8534 s & 42671 & 42 & 790 & 171 & 1200 $\times$ 900 & 390 - 950 nm & 8 \\
    \bottomrule
  \end{tabular}
\end{table*}

\subsection{Challenge Attributes}

\begin{figure*}[t]
  \centering
  \includegraphics[width=1\linewidth]{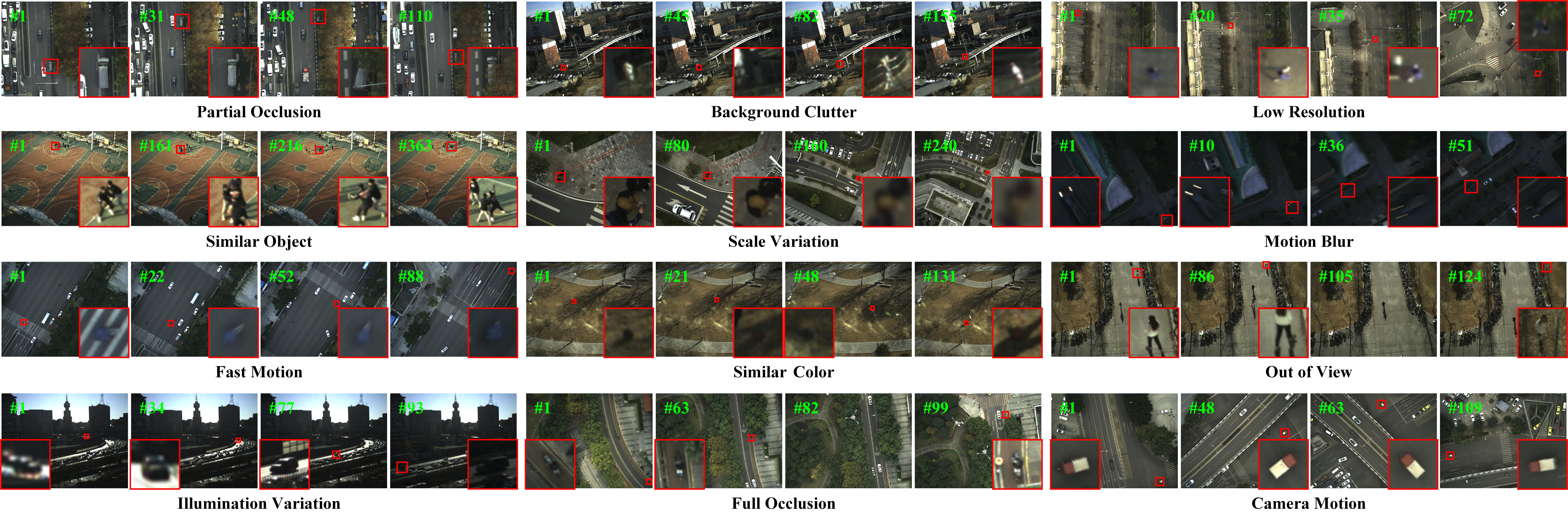}
  \caption{Typical examples of each challenge attribute in the MUST dataset, with the zoomed-in target areas.}
  \label{fig:challenge}
\end{figure*}

\begin{table}
  \caption{Detailed description of the challenge attributes.}
  \label{tab:challenge}
  \centering
  \setlength{\tabcolsep}{2pt}
  \renewcommand{\arraystretch}{1}
  \begin{tabular}{cl}
    \toprule
    Attribute & Description \\
    \midrule  
    \textbf{POC} & 
    \begin{tabular}[c]{@{}l@{}}\textbf{Partial Occlusion}: The target is partially occ-\\luded by objects.\end{tabular} \\
    \textbf{BC} & \begin{tabular}[c]{@{}l@{}}\textbf{Background Clutter}: The background of the \\ target is cluttered.\end{tabular}       \\
    \textbf{LR} & \begin{tabular}[c]{@{}l@{}}\textbf{Low Resolution}: The target area is less than \\ 100 pixels. \end{tabular}\\
    \textbf{SOB} & \begin{tabular}[c]{@{}l@{}}\textbf{Similar Object}: Similar objects exist around \\ the target. \end{tabular}\\
    \textbf{SV} & \begin{tabular}[c]{@{}l@{}}\textbf{Scale Variation}: The target changes signific-\\antly in size or shape.       \end{tabular} \\
    \textbf{MB} & \begin{tabular}[c]{@{}l@{}}\textbf{Motion Blur}: The movement of the target \\ creates blur.          \end{tabular}    \\
    \textbf{FM} & \begin{tabular}[c]{@{}l@{}}\textbf{Fast Motion}: The distance between targets in \\ two adjacent frames exceeds 20 pixels.       \end{tabular}     \\
    \textbf{SC} & \begin{tabular}[c]{@{}l@{}}\textbf{Similar Color}: The target and background are \\ visually similar in color.        \end{tabular}    \\
    \textbf{OV} & \begin{tabular}[c]{@{}l@{}}\textbf{Out of View}: The target moves out of view and \\ returns to view after a while.          \end{tabular}    \\
    \textbf{IV} & \begin{tabular}[c]{@{}l@{}}\textbf{Illumination Variation}: The illumination state \\ of the environment around the target changes.  \end{tabular} \\
    \textbf{FOC} & \begin{tabular}[c]{@{}l@{}}\textbf{Full Occlusion}: The target is completely occ-\\luded by other objects. \end{tabular} \\
    \textbf{CM} & \begin{tabular}[c]{@{}l@{}}\textbf{Camera Motion}: The UAV platform appears \\ to be shaking or rotating.    \end{tabular}     \\
    \bottomrule
  \end{tabular}
\end{table}

To comprehensively evaluate state-of-the-art trackers, our proposed MUST dataset have defined 12 key challenge attributes that reflect the unique characteristics of multispectral UAV tracking tasks. These attributes are: Partial Occlusion (POC), Background Clutter (BC), Low Resolution (LR), Similar Object (SOB), Scale Variation: (SV), Motion Blur (MB), Fast Motion (FM), Similar Color (SC), Out of view (OV), Illumination Variation (IV), Full Occlusion (FOC) and Camera Motion (CM), as summarized in \cref{tab:challenge}. Each video sequence in the dataset is annotated with multiple challenge attributes, and \cref{fig:challenge} exhibits typical examples of each attribute.

These challenge attributes present significant difficulties for traditional RGB-based trackers, whereas multispectral-based trackers offer more effective solutions. For instance, in the cases of Background Clutter and Similar Color, the target often shares similar appearance characteristics with surrounding objects and backgrounds, making them challenging to distinguish using RGB data alone. Moreover, in scenarios involving occlusion or small target sizes, the limited spatial features available in RGB images are insufficient to maintain robust tracking, leading to tracking drift. In contrast, multispectral data leverages spectral information that is inherently linked to the target's material composition, providing more stable and distinctive features throughout the tracking process. These attributes highlight the potential of multispectral data for improving UAV tracking performance in challenging scenarios.

\section{Asymmetric Attention}

\begin{figure*}
  \centering
  \includegraphics[width=0.9\linewidth]{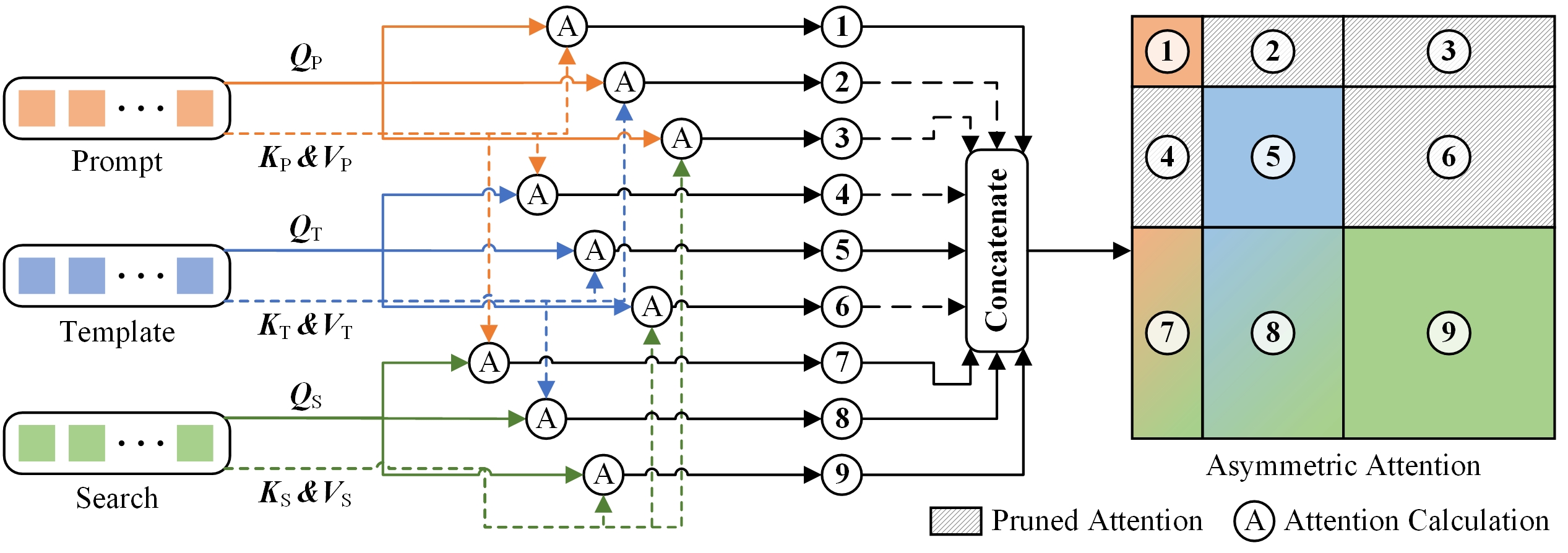}
  \caption{Illustration of asymmetric attention calculation.}
  \label{fig:ays}
\end{figure*}

Our proposed UNTrack employs an asymmetric attention structure to enable unified spectral-spatial-temporal feature extraction. The attention calculation process is illustrated in \cref{fig:ays}. Specifically, we model the relationship between embedded query $\textbf{\emph{Q}} = [\textbf{\emph{Q}}_{\rm{P}};\textbf{\emph{Q}}_{\rm{T}};\textbf{\emph{Q}}_{\rm{S}}]$, key $\textbf{\emph{K}} = [\textbf{\emph{K}}_{\rm{P}};\textbf{\emph{K}}_{\rm{T}};\textbf{\emph{K}}_{\rm{S}}]$, and value $\textbf{\emph{V}} = [\textbf{\emph{V}}_{\rm{P}};\textbf{\emph{V}}_{\rm{T}};\textbf{\emph{V}}_{\rm{S}}]$, generating the corresponding attention maps. These maps are divided into nine distinct blocks, each representing the interaction between different tokens:
\begin{enumerate}[(1)]
\item Self-attention on Prompt;
\item Cross-attention on Prompt with Template;
\item Cross-attention on Prompt with Search;
\item Cross-attention on Template with Prompt;
\item Self-attention on Template;
\item Cross-attention on Template with Search;
\item Cross-attention on Search with Prompt;
\item Cross-attention on Search with Template;
\item Self-attention on Search.
\end{enumerate}

Recent research has shown that each block plays an inconsistent role in the tracking process~\cite{cui2022mixformer, song2023compact}. For example, blocks 1, 5, and 9 correspond to the self-attention results of the prompt, template, and search, respectively. These blocks capture deep semantic information that significantly enhances tracking performance. Moreover, the core of visual object tracking lies in accurately locating the target within search regions, a process that relies on effective cross-information aggregation in block 8. Given that the prompt encodes historical spectral information of the target, block 7, representing the interaction between the search and prompt, plays a crucial role in improving target discrimination, particularly in complex backgrounds.

In contrast, blocks 2 and 3 correspond to interactions between the prompt and other tokens, which tend to introduce noise and pollute the spectral information, thereby negatively affecting tracking accuracy. Blocks 4 and 6, which focus on the template, irrelevant to the generation of tracking results and only add unnecessary computational overhead. These blocks not only fail to aid in tracking, but also increase the computational cost without improving performance. Consequently, we prune blocks 2, 3, 4, and 6 in the asymmetric attention structure, resulting in a more compact attention map that enhances tracking precision while reducing computational burden.

\begin{table}[t]
\label{tab:param}
\caption{Detailed description of the challenge attributes.}
  \centering
  \setlength{\tabcolsep}{8pt}
  \begin{tabular}{l|cc}
    \toprule
     Infrared bands & AUC ($\%$) & Pre ($\%$) \\
     \midrule
     Training from scratch & 55.9 & 74.1 \\
    Replicating & 59.7 & 79.2 \\
    \bottomrule
  \end{tabular}
\end{table}

\section{Experiment Details}

\subsection{Parameter Reconstruction}

In previous multispectral images processing, a prevalent practice adopts the input layer weights of pre-trained RGB-based networks to initialize those of multispectral-based networks~\cite{liu2022siamhyper, ying2024hyperspectral}. However, this approach ignores the differences between spectral bands, without considering the prior information from the spectral camera. In this work, we propose a simple yet effective parameter reconstruction strategy that enables the use of RGB-based pre-trained parameters for multispectral vision tasks. 

\begin{figure*}
  \centering
  \includegraphics[width=1\linewidth]{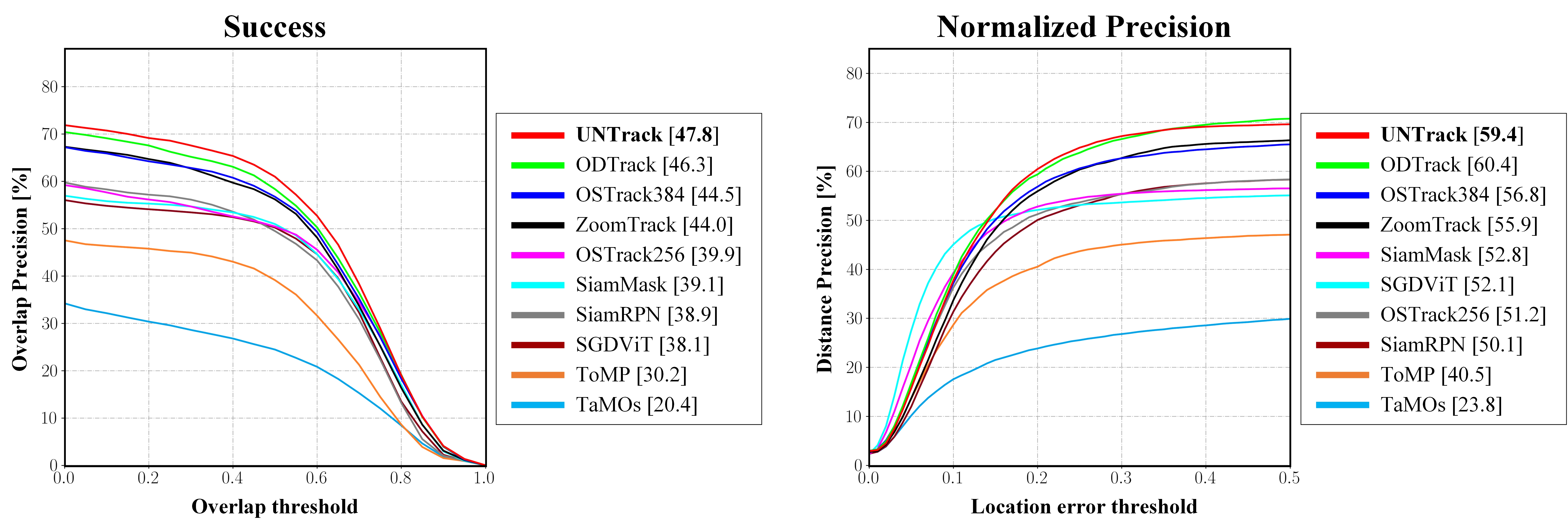}
  \caption{Success and Normalized Precision plots for state-of-the-art trackers on the MUST dataset.}
  \label{fig:curves}
\end{figure*}

\begin{figure*}
  \centering
  \includegraphics[width=1\linewidth]{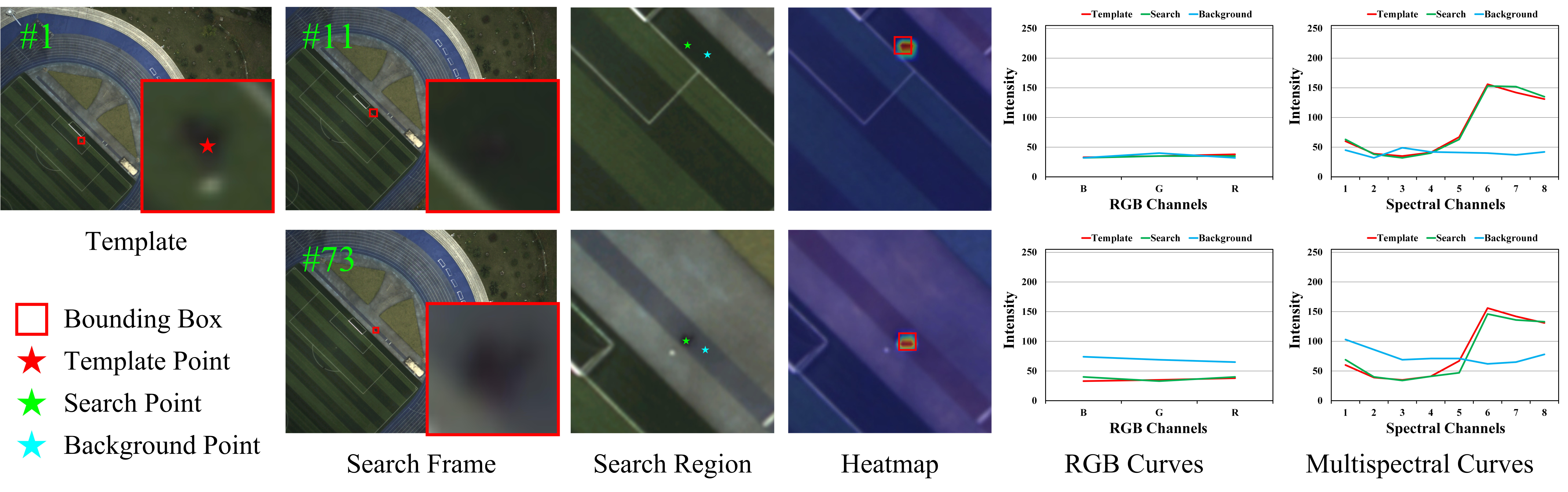}
  \caption{Performance of the proposed UNTrack in scenarios with similar target and background colors. We sample points for template, search, and background during tracking and visualize their spectral curves.}
  \label{fig:spectral}
\end{figure*}

Specifically, we assign a spectral band to each channel of the RGB-based parameters, following the definitions provided by the CIE standard~\cite{stockman2019cone}, which correspond to the wavelengths: $\rm{Red=700.0nm}$, $\rm{Green=546.1nm}$, $\rm{Blue=435.8nm}$. Following previous works, we interpolate pre-trained RGB-based weights to initialize the input layer weights corresponding to visible bands. For infrared bands, we replicate the pre-trained RGB-based red channel weights, which are then assigned as the initial weights to each infrared band. The reconstruction process can be formalized as follows:
\begin{equation}
  \textbf{\emph{W}}_{{\rm{M}}_{i}} = \left\{
  \begin{aligned}
  &\frac{\textbf{\emph{W}}_{\rm{B}}({\rm{G}} - {\rm{M}}_{i}) + \textbf{\emph{W}}_{\rm{G}}({\rm{M}}_{i} - {\rm{B}})}{{\rm{G-B}}},if\enspace{\rm{M}}_{i} < {\rm{G}} \\
  &\frac{\textbf{\emph{W}}_{\rm{G}}({\rm{R}} - {\rm{M}}_{i}) + \textbf{\emph{W}}_{\rm{R}}({\rm{M}}_{i} - {\rm{G}})}{{\rm{R-G}}},if\enspace{{\rm{G}}<\rm{M}}_{i}\leq {\rm{R}} \\
  &\textbf{\emph{W}}_{\rm{R}},if\enspace{{\rm{R}}<\rm{M}}_{i} \\
  \end{aligned}
\right.
  \label{eq:reconstruction}
\end{equation}
where $\rm{R,G,B}$ and $\textbf{\emph{W}}_{\rm{R}}, \textbf{\emph{W}}_{\rm{G}}, \textbf{\emph{W}}_{\rm{B}}$ represent the spectral bands and weights corresponding to the red, green, and blue channels in the RGB space. ${\rm{M}}_{i}$ and $\textbf{\emph{W}}_{{\rm{M}}_{i}}$ denote the $i$-th spectral band and the corresponding channel weights in MUST, where $i=1,\dots,8$.

Notably, we compared the performance of different infrared bands initialization approaches. As shown in \cref{tab:param}, we evaluate initializing infrared bands by training from scratch instead of replicating. Without utilizing pre-trained RGB-based weights as prior information, training infrared bands from scratch reduces AUC by $3.8\%$.

Our proposed initialization strategy is universally applicable, for example, it improves AUC of OSTrack$_{256}$ by $12.4\%$, UNTrack by $11.9\%$ and ZoomTrack by $9.3\%$. The exception is ODTrack, Which embeds video-clip input as the template but neglects spectral redundancy, resulting in limited improvement. ODTrack's inherent flaw leads to a shift in suboptimal methods, from ODTrack to OSTrack$_{384}$, before and after initialization. Consequently, UNTrack has performance discrepancy when compared with two suboptimal methods before and after initialization.

\subsection{Comparative Analysis}

To further evaluate the performance of our UNTrack against state-of-the-art trackers, we present the Success Plot and Normalized Precision Plot on the MUST dataset in \cref{fig:curves}. As shown, UNTrack outperforms the runner-up, OSTrack~\cite{ye2022joint}, with a performance gain of $1.5\%$ in success rate, and demonstrates superior performance compared to other trackers. This improvement is attributed to UNTrack's comprehensive utilization of spectral information, which enables it to effectively address complex tracking challenges.

\subsection{Visualization Analysis}
To highlight the potential of multispectral-based tracking, we visualize UNTrack’s performance on a challenging scenario characterized by the Similar Colors challenge attribute. As shown in \cref{fig:spectral}, when the target and background exhibit similar colors, traditional RGB-based data fail to distinguish them due to the lack of intensity differences across the RGB channels, leading to poor performance. In contrast, the spectral curves of the target and background exhibit distinct differences, with the target’s spectral curve remaining stable throughout the tracking process. Leveraging this characteristic, UNTrack maintains accurate tracking by focusing on the target, even in situations where color similarity complicates tracking in the RGB space.
{
    \small
    \bibliographystyle{ieeenat_fullname}
    \bibliography{main}
}


\end{document}